\documentclass[final]{cvpr}

\usepackage{times}
\usepackage{epsfig}
\usepackage{graphicx}
\usepackage{amsmath}
\usepackage{amssymb}
\usepackage{marvosym}
\usepackage{booktabs}

\usepackage[ruled,vlined]{algorithm2e}
\newcommand{\dtoprule}{\specialrule{1pt}{0pt}{0.4pt}%
            \specialrule{0.3pt}{0pt}{\belowrulesep}%
            }

\begin{document}

\title{SB-MTL: Score-based Meta Transfer-Learning for Cross-Domain \\ Few-Shot Learning }

\author{John Cai \\
Princeton University \\
  {\tt jjcai@alumni.princeton.edu}
\and
Bill Cai \\
Massachusetts Institute of Technology \\
  {\tt billcai@alum.mit.edu} 
\and 
Shen Sheng Mei \\
Pensees Pte Ltd \\
  {\tt jane.shen@pensees.ai}
}

\maketitle

\begin{abstract}
  While many deep learning methods have seen significant success in tackling the problem of domain adaptation and few-shot learning separately, far fewer methods are able to jointly tackle both problems in Cross-Domain Few-Shot Learning (CD-FSL). This problem is exacerbated under sharp domain shifts that typify common computer vision applications. In this paper, we present a novel, flexible and effective method to address the CD-FSL problem. Our method, called Score-based Meta Transfer-Learning (SB-MTL), combines transfer-learning and meta-learning by using a MAML-optimized feature encoder and a score-based Graph Neural Network. First, we have a feature encoder with specific layers designed to be fine-tuned. To do so, we apply a first-order MAML algorithm to find good initializations. Second, instead of directly taking the classification scores after fine-tuning, we interpret the scores as coordinates by mapping the pre-softmax classification scores onto a metric space. Subsequently, we apply a Graph Neural Network to propagate label information from the support set to the query set in our score-based metric space. We test our model on the Broader Study of Cross-Domain Few-Shot Learning (BSCD-FSL) benchmark, which includes a range of target domains with highly varying dissimilarity to the miniImagenet source domain. We observe significant improvements in accuracy across 5, 20 and 50 shot, and on the four target domains. In terms of average accuracy, our model outperforms previous transfer-learning methods by 5.93\% and previous meta-learning methods by 14.28\%.
\end{abstract}

\section{Introduction}

In the past decade, vast improvements to visual recognition systems have been achieved by training deep neural networks on ever-expanding training data-sets \cite{krizhevsky2017imagenet} \cite{he2016deep}. However, the performance of deep neural networks has directly depended on the size and variance of the training data-sets \cite{recht2019imagenet}. Unfortunately, acquiring large and representative data-sets is extremely costly in both time and resources \cite{deng2009imagenet}. Furthermore, when dealing with rare examples in medical (e.g. rare cancers) or satellite images (e.g. oil spills), our ability to obtain labelled samples is limited.

To address the limitations of previous deep learning methods, few-shot learning (FSL) methods have been developed to explicitly optimize models that predict new classes using only a few labelled images per class. More recent FSL methods have displayed performance converging to that of typical supervised deep learning \cite{wang2020generalizing}.

However, existing few-shot learning methods fail to generalize when there is domain-shift, as these methods have been developed with the assumption that the training and test data-sets are drawn from the same distribution \cite{chen2019closer}. This places a significant restriction on the utility of existing few-shot learning methods, as domain-shifts are common in practical applications \cite{venkateswara2017deep}. Furthermore, new domains arise frequently in computer vision, and having a robust cross-domain few-shot learning solution would allow new domains to be quickly conquered \cite{wilson2020survey}.

Hence, the Broader Study of Cross-Domain Few-Shot Learning (BSCD-FSL) was introduced by \cite{guo2020broader} as a new benchmark to test for generalization ability across a range of vastly dissimilar domains. The benchmark includes domains from natural, satellite and medical images, which has a range of perspective and color. This is in contrast to previous attempts at considering Cross-Domain Few-shot Learning \cite{chen2019closer} \cite{tseng2019cross} \cite{triantafillou2019meta}, as these attempts were limited to natural images. Moreover, the domains in BSCD-FSL are also common applications of computer vision today.

Critically, however, the results in BSCD-FSL show that current state-of-the-art meta-learning methods significantly underperform compared to transfer-learning methods \cite{guo2020broader}. This is likely due to the inability of meta-learning to transfer features from the source domain to the target domain when the domain shift is large like those in BSCD-FSL.

Hence, the main objective of this paper is to propose a meta-learning method that performs well on sharp domain shifts found in the BSCD-FSL benchamrk. To this end, we integrate the best aspects of transfer-learning with the best aspects of meta-learning to learn how to transfer. Thus, we propose a novel method that is conceptually simple yet effective, which is Score-based Meta-Transfer Learning.

At the initial stage of Score-based Meta Transfer-Learning, we include a deep residual network \cite{he2016deep} that has the last few layers designed to be fine-tuned. This is done using a modified first-order MAML-based algorithm \cite{nichol2018first}. Subsequently, the predicted scores (which correspond to pre-softmax probabilities) from the fine-tuning process are linearly mapped onto a metric space through a learned linear layer. A Graph Neural Network (GNN) then predicts a label by aggregating the message propagated from the support set to the query image \cite{satorras2018few}. Each message contains information on actual labels and initial predicted scores.

During test time, we also use data augmentation and further boost the accuracy derived from fine-tuning. We also demonstrate how the addition of the graph neural network improves performance over merely using a first-order MAML-based algorithm. In our final results, we demonstrate the superiority of our proposed system over both existing transfer-learning and meta-learning methods \cite{guo2020broader}.

The main objective behind using classification scores as the input of the GNN is to address the problem of overfitting of the metric space. By having a metric based on classification scores, we create a metric space that only considers the relations between classification scores. These score-based relations are domain-agnostic as they not directly related to domain-specific features. The GNN also learns how to re-interpret patterns in the initial classification scores to correct for asymmetric confusion \cite{DIEZ20181}.

 Our key contributions can be summarized as:  

\begin{enumerate}
    \item We achieve state-of-the-art performance on BSCD-FSL over existing meta-learning and transfer-learning methods. In some settings, our methods achieve accuracies comparable to typical supervised deep learning.
    
    \item We provide a simple and effective method to combine transfer-based (learning to fine-tune) and metric-based meta-learning methods.
    \item We are the first to propose using classification scores to create a metric space for re-classification by a Graph Neural Network. We demonstrate its effectiveness and highlight reasons why it works well. 
\end{enumerate}
\begin{figure*}[t]
\begin{center}
   \includegraphics[width=0.87\linewidth]{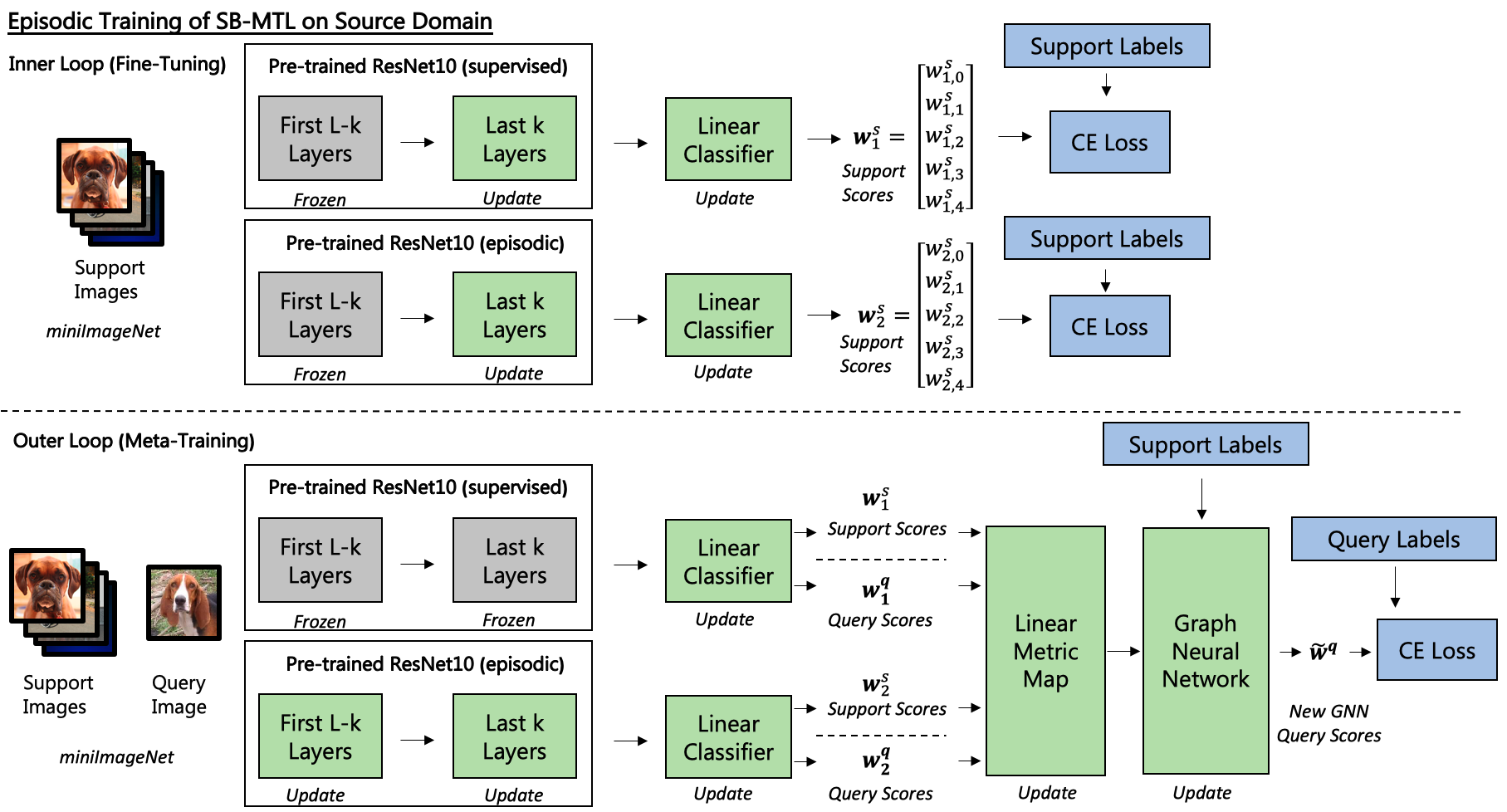}
\end{center}
   \caption{Episodic training for our Proposed Score-based Meta Transfer-Learning Model (SB-MTL)}
\label{fig:SBMTL}
\end{figure*}
\section{Cross-Domain Few-Shot Learning}

A few-shot learning problem is formalized as follows. We have a \(N_w\) way (number of classes) and \(N_s\) shot (number of support samples). Denote the collection of input images as \(\mathcal{X} = (X_0, X_1, ... , X_n)\) and the class labels as \(\mathcal{Y} = (Y_0, Y_1, ..., Y_n)\).

An iteration during the few-shot learning problem is an episode where we sample \(N_w\) classes and construct a task \(T\). This task \(T\) consists of a support set \(\mathcal{S}_s = \{(\mathcal{X}_s, \mathcal{Y}_s\})\) with  \( |\mathcal{S}_s| =N_w \times N_s\) and a query set \(\mathcal{S}_q = \{(\mathcal{X}_q, \mathcal{Y}_q)\}\) with \( |\mathcal{S}_q| =N_w \times N_q\). A model is typically adapted on the support set and then tested on the query set. 

Denote a domain as a joint distribution \(P\) over the input image space and label space \(\{(\mathcal{X}, \mathcal{Y})\}\). With the addition of the cross-domain setting we have the following: a source domain \(\{(\mathcal{X}_\alpha, \mathcal{Y}_\alpha)\}\) and a target domain source domain \(\{(\mathcal{X}_\beta, \mathcal{Y}_\beta)\}\), where \(P_\alpha \neq P_\beta\) and label spaces are disjoint \(\mathcal{Y}_\alpha \cap \mathcal{Y}_\beta = \varnothing\) \cite{guo2020broader}.

Cross-domain few-shot learning models are typically trained in an episodic manner: tasks \((T_0, T_1,...,T_n)\) are drawn from the source domain, with the model adapted to each support set, and the loss calculated over each query set.

\section{Relevant Work}

\subsection{Few-Shot Learning}
There are three key sets of methods used in few-shot learning: metric-based methods, transfer-based methods and data augmentation. 

Metric-based methods aim to learn a metric function \(\phi_m\) that can be used to classify query images based on their distances from the few support images. The approach can be generalized in the following way \cite{tseng2019cross}. 

Let the feature encoder be \(\phi_f\). Let a query image's feature vector be \(\phi_f(X_q) \in \mathbb{R}^d\) and let the support image set's feature vectors form the matrix \(\phi_f(X_S) \in \mathbb{R}^{N_s \times d}\). Then, 
\begin{equation}
    \hat{Y}_q = \phi_m (Y_s, \phi_f(X_S), \phi_f(X_q))
    \label{eq: phim}
\end{equation}
The training objective is cross-entropy loss \(L(Y_q,\hat{Y}_q)\).

A popular metric-based method is prototypical networks \cite{snell2017prototypical}, which generates mean-centroids ``prototypes" using support examples. Further classification of query examples is performed using a nearest centroid approach in Euclidean space. A natural extension of this method introduces a Relation Network \cite{sung2018learning} that learns a deep distance metric to compare the feature vectors. Other metric learning methods include learning embeddings that can generalize under a linear support vector machine \cite{lee2019meta}.

The key metric-based method that we use is \textbf{Graph Neural Network}. Graph-based convolutions can create more flexible representations of data, which can be a benefit when there is no inherent reason to assume an Euclidean structure to the data \cite{bronstein2017geometric}. Moreover, \cite{satorras2018few} shows that previous metric-based methods such as Protypical Networks and Relation Networks have corresponding graph representations with fixed weights. We expand on this key method later.

Transfer learning involves reusing features learned from base classes \cite{pan2009survey}. This is typically performed by fine-tuning a pre-trained model to perform well on the unseen classes. Transdusctive fine-tuning uses batch normalization \cite{ioffe2015batch} statistics of all samples in the episode for prediction rather than individually predicting the label for each query sample independently. Surprisingly, applying transdusctive fine-tuning on a simple pre-trained model has comparable performance to previous metric-learning methods \cite{dhillon2019baseline}. 

A simple extension of fine-tuning would be to \textbf{learn to fine-tune}. Methods such as Model Agnostic Meta Learner (MAML) \cite{finn2017model} learn an internal representation that can be fine-tuned on new tasks using a few gradient steps. First-order approximations to the MAML, such as Reptile \cite{nichol2018first}, trains faster but perform as well on established benchmarks. Other more complex methods include a few-shot optimisation approach that applies an LSTM-based meta-learner model to learn the optimization strategies used to train another learner network \cite{ravi2016optimization}. We apply the first-order MAML algorithm, together with transdusctive fine-tuning, to learn how to fine-tune our feature encoder.

Data augmentation generates more images from the sparse support images to improve generalization. Methods range from mechanical augmentations such as rotation, jitter and crops \cite{krizhevsky2017imagenet} to using generative adversarial networks to learn complex augmentations \cite{antoniou2017data} \cite{wang2018low}. Despite the obvious benefits of data augmentation for overcoming domain shifts and sparse samples, augmentation during fine-tuning has not been  applied to previous work in CD-FSL. 

\subsection{Domain Adaptation}

Most previous work on domain adaptation has been in a context where the training and test label spaces are aligned \(\mathcal{Y}_\alpha = \mathcal{Y}_\beta \). This means that the training and test data contain the same set of classes, but have a different statistical distribution of features \cite{weiss2016survey}. An example would be adapting a model from photos to drawings of a person's face \cite{wang2018deep}.  However, when labels spaces are disjoint, as in the CD-FSL problem, class-based alignment methods cannot be directly applied and the problem is more difficult to solve. \cite{weiss2016survey}.

\subsection{Cross-Domain Few-Shot Learning}

Most previous work on CD-FSL has domains only belonging to natural images \cite{chen2019closer} \cite{koniusz2018museum} \cite{tseng2019cross}, with two notable exceptions: \cite{guo2020broader} and \cite{zou2020revisiting}. In, \cite{zou2020revisiting}, they combine mid-level features from the neural network using a residual prediction task to tackle the CDFSL problem when transferring from ImageNet to medical malaria images.

In \cite{tseng2019cross}, simple learned Feature-Wise Transformations (FWT) are combined with meta-learning methods to aid domain adaptation. In \cite{guo2020broader}, all previous meta-learning methods, inclusive of the use of feature-wise transformations, under-perform by 12.8\% even when compared to simple fine-tuning methods.  In contrast, transfer-learning methods such as Transductive Fine-tuning (TransFT) \cite{dhillon2019baseline} and Incremental Multi-Model Selection (IMS-f) perform significantly better. IMS-f, however, transfers from multiple models trained on a range of source datasets by using a linear classifier on feature vectors from each layer of different models. In our work, we restrict our pre-trained models to be trained on the miniImageNet dataset. We also differ in that we aim to train the network to learn a good initialization to fine-tune the last residual block.

\section{Methodology}

Our Score-based Meta Transfer-Learning method is shown in Figure \ref{fig:SBMTL} and comprises of three key components: 

\begin{enumerate}
    \item Meta Transfer-Learning: learning to fine-tune the feature encoder using a first-order MAML algorithm
    \item Score-based Metric Learning: Mapping the pre-softmax classification scores of multiple models onto a metric space.
    \item Graph Neural Network: Formulation of the few-shot setting as a partially observed graphical model for re-classification in the score-based metric space.
\end{enumerate}

\subsection{Model Backbone}

To be consistent with other results in \cite{guo2020broader}, we use a ResNet10 model as the feature encoder for all our experiments. Residual networks are an effective way of allowing for feature reuse from earlier layers of the network \cite{he2016deep}. We do not use deeper ResNet models as deeper networks appears to underperform in the few-shot setting due to a higher propensity to overfit \cite{sun2019meta}. We pre-train the feature encoder on miniImageNet either through a linear classifier or a graph neural network attached to the end. These pre-trained models correspond to the baseline model in \cite{guo2020broader} and the Graph Neural Network model in \cite{tseng2019cross}.

\subsection{Meta Transfer-Learning}

After obtaining the pre-trained models, we begin the training process for meta transfer-learning outlined in Figure \ref{fig:SBMTL}. We apply a first-order MAML algorithm \cite{nichol2018first} to find a set of weight initializations for the last Residual Network block. This ensures that when faced with new examples, the last residual network block easily adapts to the new examples \cite{finn2017model}. Even though we train only on the miniImageNet source domain, the miniImageNet dataset has been found to support the training of transferable features \cite{huh2016makes}. Moreover, during the process of episodic training whereby tasks \((T_0, T_1, ..., T_n)\) are sampled from the source dataset, the distribution of images and labels mostly differ \(P_{T_i} \neq P_{T_j}\) as different classes are drawn randomly each time. Thus, there are tasks that have disjoint labels and different image distributions that simulate transfer-learning during episodic training (hence why ``Meta Transfer-Learning"). The specific training process is outlined in Algorithm \ref{algo:MTL}.

\begin{algorithm}[h]
\SetAlgoLined
 Initialize or load weights \(\phi_f\) for feature encoder, \(\phi_m\) for metric-learning module,  \(\phi_c\) for classifier and \(\phi_l\) for metric layer ; \\ 
 \For{each episode}{
  Sample \(N_s\) support samples and \(N_q\) query samples \\  
  Freeze first \(L-k\) layers of feature encoder \\ 
  \For{step = 1,2,...,}{
    Sample batch \(b\) from the \(N_s\) support samples \\
    Compute cross-entropy loss \(L_s(\phi_f, \phi_c)\) on batch \(b\).
    Update the last \(k\) layers of feature encoder \(\phi_f\) and the weights of the classifier \(\phi_c\) using SGD or Adam
  }
  
  Obtain \(\Tilde{\phi}_{f(k)}\) = \(U_b^S(\phi)\), the updated weights for last \(k\) layers \\ 
  Combine \(\phi_{f(L-k)}\) and \(\Tilde{\phi}_{f(k)}\) to obtain new feature encoder \(\Tilde{\phi}_f\)\\
  Feed images through fine-tuned feature encoder and fine-tuned classifier. For image \(X\), this gives \(\Tilde{\phi}_c(\Tilde{\phi}_f (X)) \). \(\phi_c(\Tilde{\phi}_f (X)) \) is a \(N_c\)-dimensional vector of pre-softmax classification scores \\
  Map the scores \(\phi_c(\Tilde{\phi}_f (X)) \) onto the metric space using the linear metric layer \(\phi_l\) and then through the metric learning module \(\phi_m \). \\
  Compute the cross-entropy loss on the \(N_q\) query samples and compute updates \(g_{f(L-k)}\), \(g_{f(k)}\), \(g_m\) for all model parameters using Adam \\
  Update initial parameters using learning rate \(\theta \)
 }
 \caption{Meta Transfer-Learning Algorithm}
 \label{algo:MTL}
\end{algorithm}

To compute the losses in the inner and the outer loops of the algorithm, we use the standard cross-entropy loss. The algorithm is model-agnostic and can be used with any existing metric-learning module.  The method can also be applied to a model of any backbone depth, with any number of layers set to be frozen.

\subsection{Score-based Metric Learning}
The key difference between our method and a typical MAML approach is that we do not directly use the output from our linear classifier as the final prediction scores.

Formally,  let \(X_i\) be an image.  After fine-tuning on the support set, we apply the fine-tuned feature encoder \(\Tilde{\phi}_f\) on \(X_i\) to obtain a feature vector \(\Tilde{\phi}_f (X_i) \in \mathbb{R}^{512}\).

Then, we take the fine-tuned linear classifier \(\Tilde{\phi}_c \) to produce a pre-softmax score vector in \(\mathbb{R}^5\), which corresponds to the 5-classes. Denote this score vector as \(\Tilde{F}_c(X_i) = \Tilde{\phi}_c (\Tilde{\phi}_f (X_i)) \) A learned linear layer is then applied \(\phi_l (\Tilde{F}_c(X_i)) \in \mathbb{R}^{32} \) as a metric mapping. This map the scores onto a metric space that can be used to compare support and query samples.

From previous research, we find that models trained or fine-tuned differently on the same data-set have different patterns of classification scores \cite{dvornik2019diversity}. Hence, we concatenate scores from models pre-trained in an episodic and supervised manner, and fine-tuned in different ways (simple fine-tuning and meta transfer-learning).

\begin{figure}[t]
\begin{center}
   \includegraphics[width=0.99\linewidth]{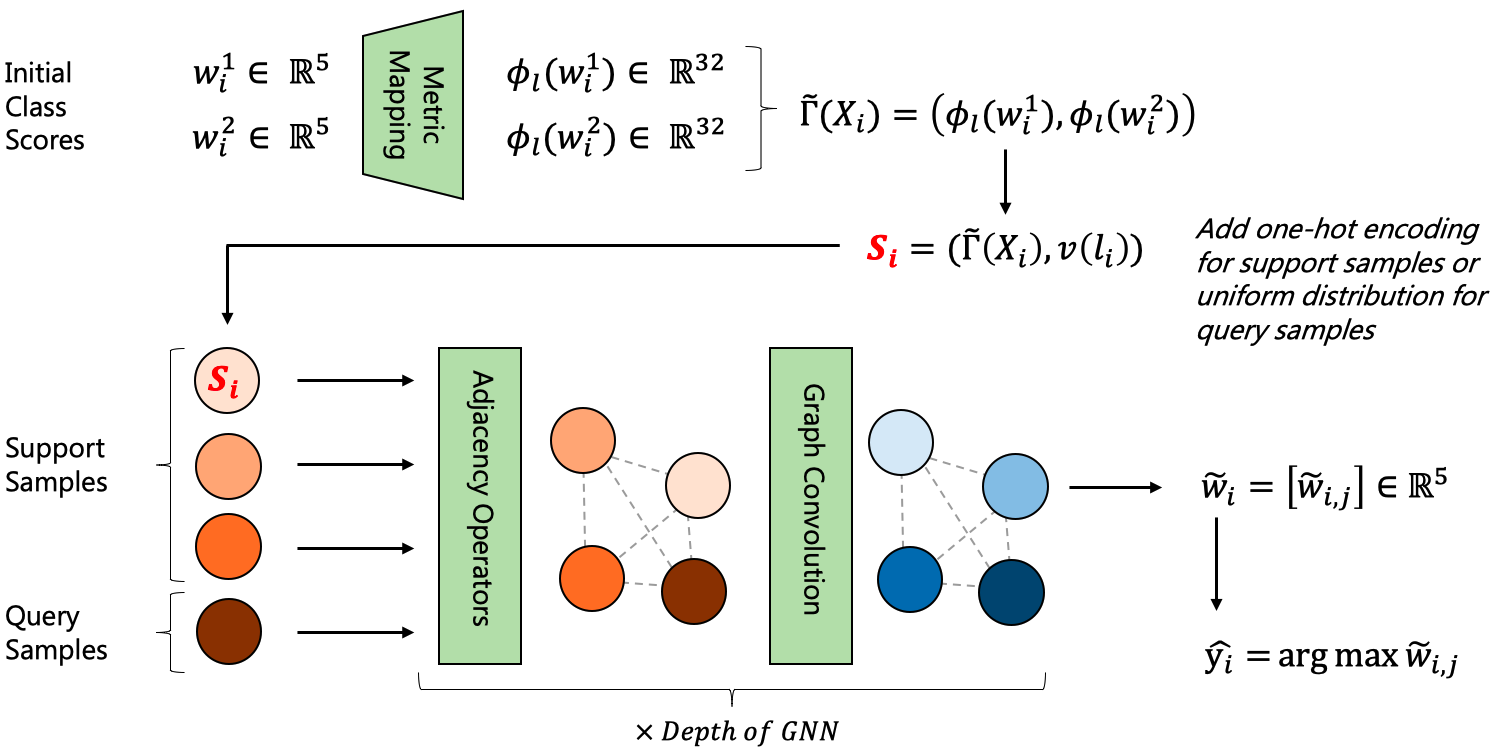}
\end{center}
   \caption{Score-based Metric Learning and subsequent Graph Neural Network (GNN)}
\label{fig:GNN}
\end{figure}

As shown in Figure \ref{fig:GNN}, we have the meta transfer-learning scores \(w^1_i = \Tilde{F}^1_c(X_i)\) and the simple fine-tuned scores \(w^2_i = F^2_c(X_i)\). We apply the same linear function on both of them to obtain \( \phi_l (w^1_i)\) and  \( \phi_l (w^2_i))\). Concatenating them gives us the following metric vector:
\begin{equation}
    \Tilde{\Gamma}(X_i) = \Big (\phi_l (w_i^1)), \phi_l (w_i^2) \Big ) \in \mathbb{R}^{64} 
\end{equation}

\subsection{Graph Neural Network}
The Meta-Learning module we use is the Graph Neural Network (GNN). We follow the formulation of the GNN for the few-shot problem in \cite{satorras2018few}. In brief, a GNN acts on local operators of a graph \(G = (V,E)\), which for the few-shot learning case is fully connected. Given the metric space vector in dimension \(d_k = 64\), the GNN takes in the input signal \(S \in \mathbb{R}^{(N_s + 1) \times d_k } \). \(N_s\) is the number of support samples, with 1 vertex for the query sample. Subsequently, a graph convolution layer \(GC(.)\) \cite{satorras2018few} is performed with linear operations on local signals. Formally, we have:
\begin{equation}
    GC(S^{k}) = f \Bigg (\sum_{B\in \mathcal{A}}B S^k \theta^{k}_{B,q} \Bigg ) , \;  q = d_1, ...,d_{k+1}  
    \label{eq:graphconv}
\end{equation}
where \(\theta^{k}_{B,i} \in \mathbb{R}^{d_k \times d_{k+1}} \), \(\mathcal{A}\) is a family of adjacency operators,  and \(f\) is a leaky ReLU function \cite{xu2015empirical}. At the next layer, the new signal would be \(S^{k+1} = GC(S^{k})\). 

In the few-shot learning formulation, we can learn the adjacency operator (edge features) using the current hidden vertex \cite{satorras2018few}. We apply a Multi-layer Perceptron (MLP) that takes in the absolute difference between the the output vectors of vertices in the graph \cite{kearnes2016molecular} \cite{gilmer2017neural}. Formally, we have the learned adjacency operator as:
\begin{equation}
    \Tilde{A}_{i,j}^{k} = \gamma(S_i^k, S_j^k) = MLP(|S_i^k - S_j^k |)
    \label{eq:edge}
\end{equation}

As shown on Figure \ref{fig:GNN}, at each layer of the GNN, we apply Equation \ref{eq:edge} to construct the edge features from the signal. In this work, our GNN has 2 layers. We can also see that the edge features satisfy distance property of symmetry and can learn the identity property \cite{satorras2018few}. 

Subsequently, we apply Equation \ref{eq:graphconv} to perform the graph convolution. The update for the vertex features is then defined by the generator family : \(\mathcal{A} = \{\Tilde{A}^{k}, \mathbf{1} \}\).  Intuitively, this convolution aggregates all messages passed from every support sample, which propagates the label information. The relative weights for each message is then determined by the learned edge features.

To initialize vertex features, we combine the linearly transformed classification scores with a one-hot encoding. As above, denote the transformed score vector as \(\Tilde{F}_c(X)\). Denote the one-hot embedding of support samples with label \(l_i\) as \(v(l_i) \in \mathbb{R}^K \) where \(K\) is the number of classes. The vertex initialization for support samples is then: 
\begin{equation}
S_i = \Big (\Tilde{\Gamma}(X_i), v(l_i) \Big)
\end{equation}
For query samples, we replace the one-hot vector \(v(l_i)\) with a uniformly distributed vector to represent label uncertainty.

\begin{table*}[t]
\makebox[\textwidth][c]{
\resizebox{\textwidth}{!}{%
    \begin{tabular}{ccccccc}
\dtoprule
 \textbf{Methods}   & \multicolumn{3}{c}{\textbf{EuroSAT}} & \multicolumn{3}{c}{\textbf{CropDisease}} \\
\cmidrule(lr){2-4}\cmidrule(lr){5-7}
   & 5-way 5-shot  & 5-way 20-shot & 5-way 50-shot  & 5-way 5-shot  & 5-way 20-shot & 5-way 50-shot \\ \midrule
ProtoNet\textsuperscript{\Cross} & 73.29\% \(\pm\) 0.71\% & 82.27\% \(\pm\) 0.57\% & 80.48\% \(\pm\) 0.57\% & 79.72\% \(\pm\) 0.67\% & 88.15\% \(\pm\) 0.51\% & 90.81\% \(\pm\) 0.43\% \\
ProtoNet+FWT\textsuperscript{\Cross} & 67.34\% \(\pm\) 0.76\% & 75.74\% \(\pm\) 0.70\% & 78.64\% \(\pm\) 0.57\% & 72.72\% \(\pm\) 0.70\% & 85.82\% \(\pm\) 0.51\% & 87.17\% \(\pm\) 0.50\%  \\
MetaOpt\textsuperscript{\Cross} & 64.44\% \(\pm\) 0.73\% & 79.19\% \(\pm\) 0.62\% & 83.62\% \(\pm\) 0.58\% & 68.41\% \(\pm\) 0.73\% & 82.89\% \(\pm\) 0.54\% & 91.76\% \(\pm\) 0.38\%  \\
FT Last1\textsuperscript{\Cross} & 80.45\% \(\pm\) 0.54\% & 87.92\% \(\pm\) 0.44\% & 91.41\% \(\pm\) 0.46\% & 88.72\% \(\pm\) 0.53\% & 95.76\% \(\pm\) 0.65\% & 97.87\% \(\pm\) 0.48\%  \\
TransFT\textsuperscript{\Cross} & 81.76\% \(\pm\) 0.48\% & 87.97\% \(\pm\) 0.42\% & 92.00\% \(\pm\) 0.56\% &  90.64\% \(\pm\) 0.54\% & 95.91\% \(\pm\) 0.72\% & 97.48\% \(\pm\) 0.56\% \\
SB-MTL & \textbf{85.93\% \(\pm\) 0.68\%} & \textbf{95.18\% \(\pm\) 0.35\%} & \textbf{97.73\% \(\pm\) 0.25\%} & \textbf{95.03\% \(\pm\) 0.42\%} &\textbf{ 99.19\% \(\pm\) 0.14\%} & \textbf{99.75\% \(\pm\) 0.08\%}  \\
\toprule
TransFT (Aug) & 82.80\% \(\pm\) 0.60\% & 90.38\% \(\pm\) 0.67\% & 93.48\% \(\pm\) 0.57\% & 92.51\% \(\pm\) 0.84\% & 97.33\% \(\pm\) 0.43\% & 98.40\% \(\pm\) 0.41\%  \\

SB-MTL (Aug) & \textbf{87.30\% \(\pm\) 0.68\%} & \textbf{96.53\% \(\pm\) 0.28}\% & \textbf{98.37\% \(\pm\) 0.18\%} & \textbf{96.01\% \(\pm\) 0.40\%} & \textbf{99.61\% \(\pm\) 0.09\%} & \textbf{99.85\% \(\pm\) 0.06\%}  \\
\bottomrule
\end{tabular}}%
}
\\[1ex]
\makebox[\textwidth][c]{
\resizebox{\textwidth}{!}{%
    \begin{tabular}{ccccccc}
\dtoprule
 \textbf{Methods}   & \multicolumn{3}{c}{\textbf{ChestX}} & \multicolumn{3}{c}{\textbf{ISIC}} \\
\cmidrule(lr){2-4}\cmidrule(lr){5-7}
   & 5-way 5-shot  & 5-way 20-shot & 5-way 50-shot  & 5-way 5-shot  & 5-way 20-shot & 5-way 50-shot \\ \midrule
ProtoNet\textsuperscript{\Cross} & 24.05\% \(\pm\) 1.01\% & 28.21\% \(\pm\) 1.15\% & 29.32\% \(\pm\) 1.12\% & 39.57\% \(\pm\) 0.57\% & 49.50\% \(\pm\) 0.55\% & 51.99\% \(\pm\) 0.52\% \\
ProtoNet+FWT\textsuperscript{\Cross} & 23.77\% \(\pm\) 0.42\% & 26.87\% \(\pm\) 0.43\% & 30.12\% \(\pm\) 0.46\% & 38.87\% \(\pm\) 0.52\% & 43.78\% \(\pm\) 0.47\% & 49.84\% \(\pm\) 0.51\%  \\
MetaOpt\textsuperscript{\Cross} & 22.53\% \(\pm\) 0.91\% & 25.53\% \(\pm\) 1.02\% & 29.35\% \(\pm\) 0.99\% & 36.28\% \(\pm\) 0.50\% & 49.42\% \(\pm\) 0.60\% & 54.80\% \(\pm\) 0.54\%  \\
FT Last1\textsuperscript{\Cross} & 25.96\% \(\pm\) 0.46\% & 31.63\% \(\pm\) 0.49\% & 37.03\% \(\pm\) 0.50\% & 47.20\% \(\pm\) 0.45\% & 59.95\% \(\pm\) 0.45\% & 65.04\% \(\pm\) 0.47\%  \\
TransFT\textsuperscript{\Cross} & \textbf{26.09\% \(\pm\) 0.96\%} & 31.01\% \(\pm\) 0.59\% & 36.79\% \(\pm\) 0.53\% & 49.68\% \(\pm\) 0.36\% & 61.09\% \(\pm\) 0.44\% & 67.20\% \(\pm\) 0.59\%  \\
SB-MTL & 25.99\% \(\pm\) 0.50\% & \textbf{33.47\% \(\pm\) 0.54\%} & \textbf{38.37\% \(\pm\) 0.56\%} & \textbf{50.68\%  \(\pm\) 0.76\%} & \textbf{68.58\% \(\pm\) 0.70\%} & \textbf{75.55\% \(\pm\) 0.58\%}  \\ 
\toprule
TransFT (Aug) & \textbf{29.23\% \(\pm\) 0.46\%} & 36.25\% \(\pm\) 0.55\% & 40.69\% \(\pm\) 0.56\% & 51.54\% \(\pm\) 0.64\% & 62.72\% \(\pm\) 0.62\% & 69.68\% \(\pm\) 0.59\%  \\

SB-MTL (Aug) & 28.08\% \(\pm\) 0.50\% & \textbf{37.70\% \(\pm\) 0.57\%} & \textbf{43.04\% \(\pm\) 0.66\%} & \textbf{53.50\% \(\pm\) 0.79\%} & \textbf{70.31\% \(\pm\) 0.72\%} & \textbf{78.41\% \(\pm\) 0.66\%}  \\
\bottomrule
\end{tabular}}%
}
\begin{footnotesize}
\Cross  -  as reported in \cite{guo2020broader}. Aug - with data augmentation during fine-tuning. \textbf{Bold} - Best performing in category. 
\end{footnotesize} 
\vskip 1mm
\caption{Results of Single Source-Domain Models on BSCD-FSL. Score-Based Meta Transfer-Learning (SB-MTL) is our proposed model. }
\label{tab:singlesource}%
\end{table*}%

\subsection{Effectiveness of Score-based GNN}

By having the pre-softmax classification scores as inputs to the GNN, we are interpreting the scores as coordinates on the metric space. Thus, we can capture more information in the patterns of the fine-tuned scores and correct a biased classifier when there is asymmetric confusion.  

To illustrate how a GNN can correct a biased classifier, let us suppose we have a fine-tuned classifier which classifies whether an image \(X\) belongs to 2 classes (A and B). Hence, the classifier outputs a vector \(w \in \mathbb{R}^2\). Suppose again that the classifier often confuses A and B, but in an asymmetric way: class \(A\) is often mistaken for class \(B\), but \(B\) is always correctly classified. This would mean that the first and second elements of the score vector \(w_0 \approx w_1\) when the classifier encounters class \(A\), but we will always have \(w_0 < w_1\)  when it encounters \(B\).  Suppose that this is the only type of error that the model makes, then the error rate on B would be zero \(P(w_0 > w_1 |B) = 0\) but the error rate on A would be non-zero \(P(w_0 < w_1 | A) > 0 \). 

By mapping the elements of \(w\) onto a metric space, the GNN can learn these relations and mitigate the confusion. All the support samples of B will be found at the coordinates corresponding to \(w_0 < w_1\), while the support samples of A will be found at the coordinates corresponding to \(w_0 \approx w_1\). The GNN's learned distance features are able to classify a new example using these coordinates: \(w_0 \approx w_1 \Rightarrow A \) and \(w_0 < w_1 \Rightarrow B\). Denote the GNN's output  as \(\Tilde{w}\) then the error rate on A will become \(P(\Tilde{w}_0 < \Tilde{w}_1 | A) = 0\).

Having sparse samples often leads to asymmetric confusion \cite{zhang2019limited} \cite{sun2020few} \cite{tsai2017improving}. The above is an elementary example of a relation in the score vector that can be learned. More complex relations can also be learned to boost accuracy.

\subsection{Data Augmentation}
For data augmentation during training, we stick to the default parameters used in \cite{guo2020broader}. For data augmentation during testing, we sample 17 additional images from the support images, and perform jitter, random crops, vertical and horizontal flips (if applicable) on a randomized basis. After fine-tuning the feature encoder, only non-augmented images are used for classification.

\subsection{Implementation Details}
As shown in Figure \ref{fig:SBMTL}, the Score-Based Meta Transfer-Learning is built on two pre-trained feature backbones. The first is a feature encoder pre-trained in an episodic manner for 400 epochs. The second is a feature encoder pre-trained in a supervised manner for 400 epochs. 

During meta transfer-learning, we learn a set of initializations for the first feature encoder. To prevent catastrophic forgetting \cite{french1999catastrophic} of features learnt during supervised training, we do not learn the second feature encoder's initializations. We still apply MAML updates to both feature encoders.

\section{Results}

\begin{table*}[t]
\makebox[\textwidth][c]{
\resizebox{\textwidth}{!}{%
    \begin{tabular}{ccccccc}
\dtoprule
 \textbf{Methods}   & \multicolumn{3}{c}{\textbf{EuroSAT}} & \multicolumn{3}{c}{\textbf{CropDisease}} \\
\cmidrule(lr){2-4}\cmidrule(lr){5-7}
   & 5-way 5-shot  & 5-way 20-shot & 5-way 50-shot  & 5-way 5-shot  & 5-way 20-shot & 5-way 50-shot \\ \midrule
All Embed \textsuperscript{\Cross} & 81.29\% \(\pm\) 0.62\% & 89.90\% \(\pm\) 0.41\% & 92.76\% \(\pm\) 0.34\% & 90.82\% \(\pm\) 0.48\% & 96.64\% \(\pm\) 0.25\% & 98.14\% \(\pm\) 0.16\%  \\
IMS-f \textsuperscript{\Cross} & 83.56\% \(\pm\) 0.59\% & 91.22\% \(\pm\) 0.38\% & 93.85\% \(\pm\) 0.30\% & 90.66\% \(\pm\) 0.48\% & 97.18\% \(\pm\) 0.24\% & 98.43\% \(\pm\) 0.16\%  \\
Linear MTL & 74.64\% \(\pm\) 0.67\% & 85.52\% \(\pm\) 0.53\% & 90.38\% \(\pm\) 0.35\% & 84.65\% \(\pm\) 0.60\% & 94.40\% \(\pm\) 0.36\% & 96.89\% \(\pm\) 0.24\%  \\
SB-MTL & 85.93\% \(\pm\) 0.68\% & 95.18\% \(\pm\) 0.35\% & 97.73\% \(\pm\) 0.25\% & 95.03\% \(\pm\) 0.42\% & 99.19\% \(\pm\) 0.14\% & 99.75\% \(\pm\) 0.08\%  \\
Linear MTL (Aug) & 77.98\% \(\pm\) 0.66\% & 89.43\% \(\pm\) 0.39\% & 93.56\% \(\pm\) 0.31\% & 88.84\% \(\pm\) 0.54\% & 97.11\% \(\pm\) 0.23\% & 98.83\% \(\pm\) 0.18\%  \\
SB-MTL (Aug) & \textbf{87.30\% \(\pm\) 0.68\%} & \textbf{96.53\% \(\pm\) 0.28}\% & \textbf{98.37\% \(\pm\) 0.18\%} & \textbf{96.01\% \(\pm\) 0.40\%} & \textbf{99.61\% \(\pm\) 0.09\%} & \textbf{99.85\% \(\pm\) 0.06\%}  \\
\bottomrule
\end{tabular}}%
}
\\[1ex]
\makebox[\textwidth][c]{
\resizebox{\textwidth}{!}{%
    \begin{tabular}{ccccccc}
\dtoprule
 \textbf{Methods}   & \multicolumn{3}{c}{\textbf{ChestX}} & \multicolumn{3}{c}{\textbf{ISIC}} \\
\cmidrule(lr){2-4}\cmidrule(lr){5-7}
   & 5-way 5-shot  & 5-way 20-shot & 5-way 50-shot  & 5-way 5-shot  & 5-way 20-shot & 5-way 50-shot \\ \midrule
All Embed \textsuperscript{\Cross}  & 26.74\% \(\pm\) 0.42\% & 32.77\% \(\pm\) 0.47\% & 38.07\% \(\pm\) 0.50\% & 46.86\% \(\pm\) 0.60\% & 58.57\% \(\pm\) 0.59\% & 66.04\% \(\pm\) 0.56\%   \\
IMS-f \textsuperscript{\Cross}  & 25.50\% \(\pm\) 0.45\% & 31.49\% \(\pm\) 0.47\% & 36.40\% \(\pm\) 0.50\% & 45.84\% \(\pm\) 0.62\% & 61.50\% \(\pm\) 0.58\% & 68.64\% \(\pm\) 0.53\%   \\
Linear MTL & 25.20\% \(\pm\) 0.43\% & 30.62\% \(\pm\) 0.45\% & 35.82\% \(\pm\) 0.47\% & 46.55\%  \(\pm\) 0.61\% &  59.14\% \(\pm\) 0.61\% & 65.35\% \(\pm\) 0.59\% \\
SB-MTL & 25.99\% \(\pm\) 0.50\% & 33.47\% \(\pm\) 0.54\% & 38.37\% \(\pm\) 0.56\% & 50.68\%  \(\pm\) 0.76\% & 68.58\% \(\pm\) 0.70\% & 75.55\% \(\pm\) 0.58\%  \\ 
Linear MTL (Aug)& 26.84\% \(\pm\) 0.44\% & 34.62\% \(\pm\) 0.48\% & 40.23\% \(\pm\) 0.56\% & 48.97\%  \(\pm\) 0.65\% & 62.99\% \(\pm\) 0.60\% & 70.32\% \(\pm\) 0.57\% \\
SB-MTL (Aug) & \textbf{28.08\% \(\pm\) 0.50\%} & \textbf{37.70\% \(\pm\) 0.57\%} & \textbf{43.04\% \(\pm\) 0.66\%} & \textbf{53.50\% \(\pm\) 0.79\%} & \textbf{70.31\% \(\pm\) 0.72\%} & \textbf{78.41\% \(\pm\) 0.66\%}  \\
\bottomrule
\end{tabular}}%
}
\begin{footnotesize}
\Cross  -  as reported in \cite{guo2020broader}. Aug - with data augmentation during fine-tuning.  \textbf{Bold} - Best performing in category. 
\end{footnotesize} 
\vskip 1mm
\caption{SB-MTL against other multi-model methods. Note that All Embed and IMS-f are multi source-domain as well. }
\label{tab:multi}%
\end{table*}%
\subsection{Experimental Setup}
Following the BSCD-FSL benchmark, we train on miniImagenet and test on CropDisease \cite{mohanty2016using}, EuroSAT \cite{helber2019eurosat}, ISIC \cite{tschandl2018ham10000} \cite{codella2019skin} and ChestX \cite{wang2017chestx} (in order of decreasing similarity). CropDisease covers plant disease, EuroSAT covers satellite images, ISIC covers dermoscopic skin lesion images and ChestX covers chest X-ray images. We report 5-way classification results for 5, 20 and 50 shot. Our code will be released at a later date.

During training, we downsample the images to speed up training. We also apply data augmentation. For pre-training, we use an Adam optimizer \cite{kingma2014adam} with 0.001 learning rate. During meta transfer-learning, we apply an Adam optimizer with 0.01 learning rate for the inner first-order MAML loop and an Adam optimizer with a 0.001 learning rate for the outer loop. In the inner loop, we also apply transductive fine-tuning. 

During meta-testing, the inner loop remains the same as that in Figure \ref{fig:SBMTL}. To get the classification scores, we use the same outer loop in Figure \ref{fig:SBMTL} and take argmax of the score-vector. We fine-tune for 15 epochs when we do not apply data augmentation, and fine-tune for 5 epochs when we apply data augmentation.

\subsection{Single Source-Domain Comparisons}

We begin by comparing our proposed model against other models trained on a single source-domain (i.e. miniImageNet). We include results from previous meta-learning methods such as ProtoNet and MetaOpt, and include results with Feature-Wise Transformation (FWT) \cite{tseng2019cross}, which fails to perform when the domain shift is large, as shown in Table \ref{tab:singlesource}. ProtoNet is the previous best-performing meta-learning method while  Transductive Fine-Tuning (TransFT) is the previous best-performing transfer-learning method \cite{guo2020broader}.

From Table \ref{tab:singlesource}, we see that even without data augmentation, SB-MTL achieves an average accuracy of \textbf{72.13\%} and outperforms all previous methods across all experimental settings with the exception of 5-way 5-shot ChestX.  In terms of average accuracy, SB-MTL outperforms TransFT by 4.00\% and ProtoNet by 12.36\%. 

With data augmentation, SB-MTL (Aug) achieves an average accuracy of \textbf{74.06\%} and outperforms all previous non-augmented methods across all experimental settings by an even larger margin. In terms of average accuracy, SB-MTL (Aug) outperforms TransFT by 5.93\% and ProtoNet by 14.28\%. To account for the effect of data augmentation, we also compare against TransFT with data augmentation. From Table \ref{tab:singlesource}, we see that SB-MTL (Aug) still significantly outperforms TransFT (Aug), with the exception of 5-way 5-shot Chest-X. In terms of average accuracy, SB-MTL (Aug) outperforms TransFT (Aug) by 1.92\%.

Another key observation is that as the number of shots increase, our outperformance over previous methods increases. In typical fine-tuning, performance increases with the number of shots because the feature encoder becomes more generalizable. In addition to the fine-tuning effect, our method also has a metric-learning effect where more labelled support samples are available, making the metric space more robust for comparing query samples to.

\subsection{Comparison to Supervised Deep Learning}

Our method is competitive with typical supervised learning on domains closer to the source domain. Typical supervised learning models for EuroSAT and CropDisease have achieved 98.57\% and 99.35\% respectively \cite{helber2019eurosat} \cite{mohanty2016using}. At our 50-shot results for SB-MTL (Aug), we achieve 98.37\% and 99.85\% respectively on EuroSAT and CropDisease. 

While there remains a considerable way before our 50-shot few-shot learning accuracies on ChestX and ISIC (43.04\% and 78.41\% ) become competitive enough with supervised deep learning, we have achieved a significant improvement (6.25\% and 13.4\%) at 50-shot for these domains over previous best-performing methods in \cite{guo2020broader} .

\subsection{Multi Source-Domain Comparisons}

From Table \ref{tab:multi}, we see that SB-MTL also outperforms previous methods that were trained on multiple source domains aside from miniImagenet. This includes All Embeddings and Incremental Multi-model Selection, which combines all embeddings from multiple pre-trained model and iteratively finds the best combination of layers respectively \cite{guo2020broader}. This outperformance is also stronger for domains closer to the source domain (EuroSAT and CropDisease).

\section{Further Analysis}

\subsection{Ablation Study: Effect of GNN}
To investigate the effect of the score-based metric space and the GNN, we include a Linear Meta Transfer-Learning (Linear MTL) model in Table \ref{tab:multi}. This model applies softmax to the initial classification scores \(w_i^1 \) and \(w_i^2\) and then adds the vectors together. Subsequently, prediction is performed simply by calling the argmax on the above vector.

From Table \ref{tab:multi}, we clearly see that there is a significant increase in accuracy when we use the score-based metric space and GNN. This gap in accuracy holds even if we were to add data augmentation to both models.  We also see that the difference in accuracy is greater for EuroSAT and CropDisease compared to ISIC and ChestX. This suggests that the GNN is more useful at boosting accuracy for target domains that are closer to the source domain.

\subsection{Asymmetry of Confusion Matrices}
As argued earlier, the score-based GNN is effective because it can learn to correct for the asymmetric confusion in the scores. A simple way to investigate this hypothesis is to compare the asymmetry in the confusion matrix of Linear MTL and our proposed SB-MTL. To this end, we introduce two measures of asymmetry. Let \(C\) be a normalized confusion matrix \(s.t. \sum_{j} (\sum_{i} C_{i,j}) = 1\). Let \(\Tilde{C} \) be the same confusion matrix with diagonal elements set to zero.
\begin{align}
    \psi_0(C) & = |C - C^T |_F \\
    \psi_1(C) & =
    \begin{cases}
   \psi_0(C) \div (2|\Tilde{C}|_F )     & \text{if }  (2|\Tilde{C}|_F ) > 0 \\
   0        & \text{otherwise}
  \end{cases}
\end{align}
where \(|.|_F\) is the Frobenius norm. The first metric is a simple way to find the norm of the matrix after subtracting the symmetric components. The second is a measure \(\in [0,1]\) that reflects what proportion of the errors in the confusion matrix are due to symmetric or asymmetric errors. At the extremes, if all errors are symmetric,    \(\psi_1(C) = 0\), and if all errors are asymmetric \(\psi_1(C) = 1\). We compare the asymmetry by sampling for 100 5-shot trials below.

\begin{table}[h]
\centering
\resizebox{\columnwidth}{!}{%
\begin{tabular}{ccccc}
\dtoprule
 \textbf{Model (Aug)}   & \textbf{CropDisease} & \textbf{EuroSAT} & \textbf{ISIC} & \textbf{ChestX} \\
  \midrule

Linear MTL  & 7.35 \(\pm\) 0.8 & 11.9 \(\pm\) 0.9  & 14.2 \(\pm\) 0.6 & 17.7 \(\pm\) 0.1 \\
SB-MTL  & 1.19 \(\pm\) 0.3 &  3.77 \(\pm\) 0.5 & 10.4 \(\pm\) 0.6 & 12.6 \(\pm\) 0.1 \\
\bottomrule

\end{tabular}}%
\label{tab:asym1}%

\caption{Asymmetry Measure \(\psi_0(C) \times 100\)  } 
\vskip 2mm
\resizebox{\columnwidth}{!}{%
\begin{tabular}{ccccc}
\dtoprule
 \textbf{Model (Aug)}   & \textbf{CropDisease} & \textbf{EuroSAT} & \textbf{ISIC} & \textbf{ChestX} \\
  \midrule
Linear MTL  & 63.95 \(\pm\) 2.0 & 58.12 \(\pm\) 2.1 & 45.6 \(\pm\) 1.6 & 42.9 \(\pm\) 1.4\\
SB-MTL  & 22.35 \(\pm\) 5.4 & 33.55 \(\pm\) 3.9  & 36.5 \(\pm\) 1.2 & 33.1 \(\pm\) 1.5\\

\bottomrule
\end{tabular}}%
\caption{Asymmetry Measure \(\psi_1(C) \times 100\)  } 
\label{tab:asym2}%
\end{table}%

As shown above, the asymmetry measures indicate that the confusion matrices of Linear MTL are much more asymmetric than SB-MTL. Moreover, the greatest difference in the asymmetry measures are found in CropDisease and EuroSAT, which is where the greatest improvements in accuracy are also made. This suggests that the GNN is learning the patterns of initial scores and correcting them.

\subsection{Benchmark Summary}

\begin{figure}[h]
\begin{center}
   \includegraphics[width=0.99\linewidth]{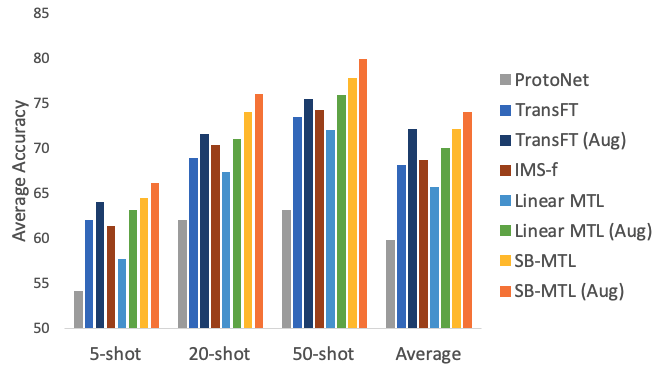}
\end{center}
   \caption{Comparison of methods across the benchmark}
\label{fig:Compare}
\end{figure}

Figure \ref{fig:Compare} summarizes the comparison of methods across the benchmark. Here, we see that our model does not suffer from the degradation in performance of typical meta-learning methods like ProtoNet in the CD-FSL setting. We also see that without the score-based GNN, Linear MTL fails to outperform previous methods. It is only with the addition of the score-based GNN that we outperform all previous models. Furthermore, while the boost from data augmentation is significant, it is far from sufficient to explain away the outperformance of our SB-MTL model.

\section{Conclusion}

In this paper, we have proposed Score-Based Meta Transfer-Learning to address the Cross-Domain Few-Shot Learning problem. To evaluate our method, we have directly tackled the difficult BSCD-FSL benchmark marked by sharp domain shifts. Our method achieves an average accuracy of \textbf{74.06\%}, which significantly outperforms previous best-performing meta-learning and transfer-learning methods by \textbf{14.28\%}  and \textbf{5.93\%} respectively.

Moreover, we achieve 50-shot accuracies on EuroSAT and CropDisease comparable to or higher than those found in typical deep supervised learning. Hence, our SB-MTL (Aug) approach could fully replace typical deep supervised learning on domains that are not too distant. With this, we can conserve considerable resources for retraining and labelling for practical cross-domain few-shot applications.

Critically, we are the first to propose the novel approach of using pre-softmax classification scores as inputs for a Graph Neural Network. We show that our score-based GNN can reclassify images by correcting the potentially biased initial scores from fine-tuning. We also provide a framework to combine MAML with a metric-learning module.

Ultimately, our work not only decisively addresses the CD-FSL problem, but it also outlines a new score-based metric learning approach that could be widely applicable for many problems in few-shot learning.

{\small
\bibliographystyle{ieee_fullname}
\bibliography{egbib}
}

\end{document}